\documentclass[11pt]{article}

\usepackage[utf8]{inputenc}
\usepackage[T1]{fontenc}
\usepackage[margin=1in]{geometry}
\usepackage{amsmath,amssymb}
\usepackage{enumitem}
\usepackage{booktabs}
\usepackage{array}
\usepackage{tabularx}
\usepackage{titlesec}
\usepackage[hidelinks]{hyperref}
\usepackage[numbers,sort&compress]{natbib}
\usepackage{xcolor}
\usepackage{graphicx}

\titleformat{\section}{\large\bfseries}{\thesection}{0.6em}{}
\titleformat{\subsection}{\normalsize\bfseries}{\thesubsection}{0.6em}{}
\setlist{itemsep=2pt,topsep=3pt}
\newcommand{\Lang}{\mathcal{L}}

\title{\textbf{Beyond Fixed Representations: The Vocabulary and Verifier Gaps in Open-Ended AI}}
\author{
  Yuan Cao\\
  \small \texttt{realcaoyuan@gmail.com}
  \and
  Haiqian Yang\\
  \small \texttt{hqyang@mit.edu}
}

\begin{document}
\maketitle

\begin{abstract}
\noindent
Modern AI systems are increasingly being evaluated for their ability to reason, code, prove theorems, use tools, and long-horizon research tasks. These are powerful capabilities, but they share a structural limitation: the representational frame within which the model operates, including its conceptual vocabulary, the space of admissible solutions it can search, and the criteria by which success is evaluated, is typically fixed and supplied in advance. This paper argues that building stronger intelligent systems capable of open-ended innovation requires additional classes of operations: the creation, stabilization, and reuse of new representational primitives, which alter the space being searched rather than simply searching within it.

We characterize the distance between current AI systems and genuinely open-ended intelligence through two gaps. The first is the vocabulary gap, the difficulty of inventing and stabilizing new representational primitives rather than merely recombining existing ones. The second is the verifier gap, the difficulty of judging the value of a new primitive when its full payoff may be visible only after future reuse. We interpret both gaps through a unified framework of intelligence as cognitive discrepancy reduction. By viewing intelligent behaviors as a sequence of cognitive transformations, we distinguish intra-space transformations which operate within a fixed representational frame, from generative transformations which may modify the frame itself. On this basis, we propose a ladder of innovation autonomy and outline several directions for advancing open-ended AI, including objectives that reward useful representational change, persistent memory architectures for invented primitives, and adaptive verification mechanisms capable of evolving alongside the representations they evaluate.
\end{abstract}

\section{Introduction}

There are two types of ``intelligent'' activities that are often conflated. The first is solving problems within a given frame: given a goal, a trained model with fixed representation space, and a way to check answers, find a good solution. The second requires changing the frame itself, for example creating new concepts, relations, measurements, abstractions, or evaluators that makes a class of problems previously unsolvable under pre-existing frames expressible and solvable. The first activity searches within a space, while the second changes the space being searched itself.

Contemporary AI excels at the first type and is still primarily benchmarked on tasks in this category. Reasoning puzzles, competition mathematics, code generation, theorem proving, tool use, and agentic task completion all evaluate performance within problems whose representational frame is largely fixed in advance---a model's weights are fixed by training data distribution; A benchmark supplies the task, the admissible form of an answer, and the standard of success; A coding environment supplies a programming language and tests; A theorem prover provides a formal language and a checker. An AI-for-science pipeline supplies a topic, a tool stack, a literature corpus, and procedures for validation. Such framed problems are useful for measuring progress, but they also impose an important limitation: our headline metrics are, by construction, weak tests of the second that require stronger capability. They tell us relatively little about whether a system can tackle genuinely open-ended innovation tasks in which the crucial step is not merely finding a solution within a given space, but recognizing that the space itself is inadequate and that a new representational primitive must be introduced.

The history of scientific and social progress can be understood, in large part, as a history of expanding representational spaces through open-ended new concept creation. Civilization advancement depends on natural, mathematical, and logical languages that progressively enlarge what can be symbolized, measured, and reasoned about. The concept of ``number'' reorganized heterogeneous collections into a common quantitative representation. The invention of ``negative numbers'' made previously impermissible algebraic operations meaningful. ``Entropy'' introduced a vocabulary for expressing regularities that were not naturally visible within older representation frameworks. Similarly, concepts such as energy, information, feedback, gene, algorithm, and attention have become representational primitives which restructure the space of what can be described, searched, measured, and reused. In the social domain, concepts such as rights, markets, incentives, and laws have played a comparable role in organizing collective behavior and political reasoning. Progress, in this sense, is not simply the accumulation of facts within a fixed framework. It is the continual transformation of the framework itself, making previously inaccessible patterns thinkable.

This is where current AI systems remain limited. Existing models are increasingly powerful at solving the type-one problems that operate within a given and frozen representational framework, where the representation primitives, search space, and evaluation criteria are largely fixed in advance. Recent analysis on LLM-driven autonomous research shows that LLM-generated research ideas are narrow and highly-concentrated on the pattern of recombining and synthesizing multiple existing ideas~\citep{tang2026airesearchagentsnarrow,chen2026measuringgaphumanllm,bao2026contemporaryailacksimagination}, which differs significantly from the human ideation patterns. To climb higher on the ladder of AGI requires systems that can also handle type-two problems: problems whose solutions require open-ended representational expansion, concept formation, and the creation of new primitives that were not supplied by the initial framework to enable autonomous innovation. In this paper, we analyze this limitation and make the following claims.

First, we argue that the distance between current AI systems and genuinely open-ended intelligent systems can be characterized by two central gaps. The first is the ``vocabulary gap'', the difficulty of inventing and stabilizing new representational primitives, rather than merely recombining existing ones supplied from outside. The second is the ``verifier gap'', the difficulty of judging whether a newly introduced primitive is worth retaining when its value is not yet validated by any criterion the system currently possesses.

Second, we interpret these innovation gaps through the lens of cognitive discrepancy reduction. Under this unified view, intelligence is driven by the pressure to reduce gaps between current states and unresolved demand states by prediction, explanation and creation. Concept invention becomes necessary when existing representations cannot sufficiently reduce such discrepancies. Intelligence, therefore, is not merely searching over known atomic units, but also transforming the space of representation so that previously unresolved tensions become tractable to narrow the gap.

Third, we define a ladder of innovation autonomy characterized by the kinds of gaps a system can close. At lower levels, systems perform \emph{intra-space} transformation, where search, reasoning, and recombination happen within a fixed representational space. At higher levels, systems must perform representational transformation that enables the invention of new primitives, operations, abstractions, and evaluation procedures that alter the representation space itself. We argue that most existing AI models and agent systems occupy the lower levels of this ladder. Moving upward requires capability augmentation beyond stronger pattern matching and completion.

Finally, we discuss what objectives and architectures would need to change in order to move AI systems toward stronger forms of open-ended intelligence. The prevalent next-token prediction objective can be interpreted as a specific instantiation of discrepancy reduction, yet this objective does not by itself reward the autonomous expansion of the system's representational space. In particular, it provides no direct mechanism for proposing, validating, stabilizing, and reusing new primitives whose value may only become apparent after the representational framework has changed. Progress toward open-ended intelligence therefore requires broader objectives and architectural designs that explicitly support representational growth.

\section{The Distance from Open-ended Intelligence}

As argued above, existing AI systems have become increasingly capable within fixed representational schemes, predefined objectives, and established problem frames. However, they remain far from stronger forms of open-ended intelligence. Such intelligence does not merely solve given problems, it also autonomously expands the space of possible problems, concepts, and methods through which new solutions can be formulated. In this section, we identify two central gaps that must be narrowed for AI systems to acquire such open-ended capabilities: the vocabulary gap and the verifier gap.

\subsection{The Vocabulary Gap}
\label{sec:vocab-gap}

Genuine discovery rests on concept invention. Before a problem can be searched, someone must supply the primitives in which candidate solutions are written, and the deepest advances are often not new answers but new primitives: a variable that did not previously exist, a relation no one had isolated, a measurement that makes a hidden regularity legible. Scientific and social progress, read this way, is in large part a history of vocabulary changes. The matrix is invented to abstract arrays of numbers and eigenvalues to capture a matrix's invariant structure. The concept of economy abstracts the production, exchange, and management of resources into an object that can be measured and studied. Current AI systems are powerful searchers and recombiners within a provided vocabulary, but they rarely decide, on their own, how and when the vocabulary should grow.

Concept creation is an act of abstraction, it extracts invariant attributes shared across a diverse range of objects. Grouping a wide range of observations under a single reusable unit shortens the descriptions of everything that uses it. This is the minimum-description-length (MDL) view of abstraction, where a primitive justifies its value by compressing a family of observations, reducing the representational budget required to express them.

In trivial cases, attaching almost any symbol to a set of raw perceptions reduces the description length and renders something newly reachable within budget. However, the critical requirement is amortization: a primitive must justify its representational value across a family of problems, not a single local instance. ``Number'' applies to any collection of objects, not only those currently being observed. ``Policy'' applies to any rule an organization might enforce, not only the actions already taken. To make this precise, let $L_{\Lang}(\cdot)$ denote description length under a language $\Lang$, and let $L^{B}_{\Lang}(f)$ denote the length of the shortest solution to task $f$ discoverable within the search budget $B$ under $\Lang$ (taken as $\infty$ if no solution is found within budget). For a task family $F$, adding a primitive $\pi$ to obtain $\Lang' = \Lang \cup \{\pi\}$ is generative with respect to $F$ and $B$ when both hold:
\begin{enumerate}[label=(\roman*)]
\item \textbf{Amortized compression:} $\displaystyle L_{\Lang}(\pi) + \sum_{f \in F} L^{B}_{\Lang'}(f) \;<\; \sum_{f \in F} L^{B}_{\Lang}(f)$ (the primitive's code cost is repaid across the family).
\item \textbf{Feasibility extension:} some class $C \subseteq F$ has $L^{B}_{\Lang}(f)=\infty$ but $L^{B}_{\Lang'}(f) < \infty$ (tasks infeasible within budget before $\pi$ become feasible afterward).
\end{enumerate}

Current models do form concepts and build vocabularies, but these are inherited from the  training distribution rather than initiated autonomously.\footnote{It is worth noting that machines can already do a bounded version of it. DreamCoder's wake–sleep cycle solves tasks, abstracts recurring program fragments into reusable library functions, and uses the enlarged library to solve future tasks. Stitch~\citep{bowers2023stitch} and related systems extract such abstractions from program corpora at scale. Nevertheless, these systems do so inside narrow regimes of program induction over formal languages, with a fixed task distribution and a supplied evaluator. The open challenge is representation expansion in general environments, across a broader range of tasks. What is more, as the next section argues, the deeper obstacle is not only inventing primitives, but knowing which ones are worth keeping.}
The abstractions a model acquires are shaped by the data distribution and limited contexts where they appear. Open-ended intelligence requires the stronger capacity to introduce primitives without that guidance. Imagine a model that has never been trained on mathematics so it has no concept of ``number''. Now show the model scenes involving objects in different quantity, like five apples, seven chairs, three people. Would it be able to autonomously create a concept of ``number'' and augment its own vocabulary and representation space for future reuse? No current system does this, and without such an autonomous vocabulary expansion, open-ended innovation is out of reach.
~\footnote{We note that concept creation need not always introduce a new symbol, but sometimes it can just assign a new stable meaning to an existing one.
This is also the kind of vocabulary growth current AI systems lack: not simply using inherited meanings, but creating or repurposing concepts that become durable tools for future thought.}

\subsection{The Verifier Gap}
\label{sec:verifier-gap}



When the representation frame is fixed, evaluation is usually fast, cheap, and decisive. A generated program passes its tests or fails, a proof is accepted by a checker or rejected, a design either improves an objective or it does not. The verifier is already defined over a fixed representation and candidate space. The strongest successes in recent AI-assisted discovery occur largely in this regime. AI discovery has advanced fastest in domains where verification is cheap, formal, and automated: code, mathematics, algorithm design, theorem proving, and simulation-heavy science. These systems became powerful primarily because the domain provides an efficient way to score candidates, making reinforcement, selection, and iterative self-improvement highly efficient. In such settings, the system can search and make recursive improvements aggressively because the target and standard of success have already been clearly supplied, so the verifier closes the loop from outside.

By contrast, verification for open-ended intelligence is much harder because verification itself becomes part of the problem. The difficulty is not just that good candidates are hard to generate, but that it may be unclear how they should be judged. There are two distinct versions of this difficulty.

The first is a delayed and expensive verification. In many real-world scientific problems, the criterion of success may be conceptually clear but practically slow or costly to observe. A biological hypothesis may require months of experiments. A drug candidate may need years of clinical validation. A materials proposal may require characterization, stress testing, and scaling-up before its value is known. In these cases, even though the ultimate verifier exists, the feedback arrives too slowly or expensively to support rapid generate-test-improve loops. This creates a serious bottleneck for autonomous discovery, but it remains largely a long-horizon credit-assignment problem where the standard of success exists, even if verification is delayed.

The second difficulty is deeper. In representation-expanding innovation, the value of a new primitive may not be measurable by the current evaluator at all. The future tasks that would justify the primitive may not yet be expressible in the current representation frame, and the standard that would validate its value may itself depend on the new representation. A new concept, relation, or measurement can change not only the space of candidate solutions but also the criteria by which answers are judged. The full value of a new representational primitive often cannot be reliably estimated by a fixed evaluator over the current unexpanded task space, because part of that value is realized only in the expanded space the primitive creates. Concepts such as entropy, information, gene, market, or algorithm did not matter simply because they optimized a pre-existing metrics. Their value was justified through the theories, experiments, institutions, and technologies that later became valuable due to the introduction of these concepts.

Closing the vocabulary gap allows a system to create new primitives, but closing the verifier gap requires a further capability: deciding which primitives are worth keeping when the current evaluator cannot yet see their full value. This is why autonomous open-ended innovation requires not only better generators, but self-extending forms of evaluation.

The verifier gap also echoes an old intuition sometimes described as the ``usefulness of the useless''\citep{flexner1939usefulness}: an idea may appear useless only because the context in which it becomes useful has not yet emerged. Modern science repeatedly illustrates this pattern. The lesson for open-ended AI is that not every valuable concept can be selected by an immediate objective or validated by a near-term outcome. Some concepts must be preserved because they expand the future space of questions, measurements, explanations, and applications. Human science partly solves this problem through distributed curiosity-driven exploration: many researchers pursue directions that are not locally optimal under current objectives, and only later does the community discover which concepts become generative. Open-ended AI may require an analogous mechanism: not merely agents that optimize fixed objectives, but exploratory populations that can retain apparently useless concepts long enough for future contexts to reveal their value.

\section{Cognitive Discrepancy Reduction}

The previous section established that a substantial distance remains between current AI and the stronger, open-ended forms of intelligence that we are after. That distance suggests a broader way of thinking about general intelligence as a whole: it is hard to call a system generally intelligent if it cannot innovate on its own, and open-ended innovation should not be treated as an independent faculty separable from reasoning, memory, planning and the rest. To study the two gaps precisely, it helps to place innovation and these other capabilities under a unified framework, and to view intelligence itself as the minimization of cognitive discrepancies.

\subsection{The Framework}
\label{sec:framework}
By a cognitive discrepancy we mean the mismatch between a system's current representational state and a desired goal state. Much of human intellectual activity can be treated as the continual effort to reduce the mismatch. To be more precise, let $R_t$ denote the system's internal representational state at time $t$, including its current stock of concepts, relations, operators, memories, tools, and hypotheses, and $G_t$ denote a target condition. Note the target need not be a fully specified state, especially in the case of concept invention where it usually cannot be, since the representation that would express the goal does not exist yet. It is better understood as a family of acceptable states: a more compressed description, a coherent explanation, a successful intervention---any condition that marks progress toward better understanding or control of the world.

We can then write a general schema for intelligence as trajectory-level discrepancy reduction, carried out through a sequence of $K$ cognitive transformations:

\begin{equation}
\label{eq:discrepancy}
\min_{{T_t}}
\sum_{t=0}^{K}
\mathcal{D}_t(R_t, G_t),
\qquad
R_{t+1}=T_t(R_t).
\end{equation}

\noindent where $\mathcal{D}_t$ measures the discrepancy between the current representational state and the target condition at step $t$. Depending on context, this discrepancy may take the form of prediction error, lack of a usable concept, an incomplete explanation, a control error, or an inconsistency. The subscript $t$ indicates that in open-ended intelligent behavior, neither the target $G_t$ nor the discrepancy measure $\mathcal{D}_t$ should be fixed once and for all. As the representation evolves, the system may discover new sub-goals, new questions, new standards of verification, until the target is achieved.

The transformations $T_t$ include a set of fundamental cognitive operations:

\begin{itemize}[leftmargin=*]
\item \emph{Association}: links related concepts and form relational structures.
\item \emph{Abstraction}: extracts invariants across cases and representing them at a higher level.
\item \emph{Analogy}: maps structures with relational alignment from a known domain onto a new one.
\item \emph{Composition}: combines existing primitives into larger conceptual, functional, or procedural structures.
\item \emph{Concept creation}: stabilizes invariants as a named, reusable primitive to make abstraction persistent.
\end{itemize}

These operations are not independent modules. Abstraction may depend on analogy across cases, analogy may depend on association between remote domains, composition may produce structures that later become candidates for abstraction, and concept creation may require several of these operations at once. The list is therefore not meant as a complete set of cognitive primitives, but as a vocabulary for describing transformations $T_t$ that reduce discrepancy. What is more, these operations have to be guided and regulated by cognitive priors. Without priors, the space of possible transformations is intractably large: infinitely many abstractions, analogies, and compositions can be proposed. Useful cognition therefore requires regularizing assumptions about what kinds of transformations are likely to matter. In human cognition these include, among others, simplicity, symmetry, spatial and temporal continuity, causal coherence, compositionality, and preference for reusable structure. Formally, one can represent these priors as a cost or preference term $\Omega(T_t, R_t)$, yielding a local objective of the form

\begin{equation}
\label{eq:transformation}
T_t \in \arg\min_{T}
\left[
\mathcal{D}_t\big(T(R_t),\, G_t\big) + \lambda\,\Omega_t(T, R_t)
\right]
\end{equation}

\noindent Again, the equation is schematic rather than algorithmic. The point is not that cognition solves a global optimization problem over all possible futures. Instead, it  that intelligent behavior can be described as a sequence of locally constrained transformations that reduce the mismatch between representational status and task demand.

This framing explains why familiar forms of intelligence look different on the surface but share a common structure. In perception, the gap is between sensory input and a stable interpretation. In prediction, it is between expected and observed events. In planning, it is between the current state and a desired future state. In explanation, it is between observed regularities and the system’s available causal or conceptual model. In analogy, it is between an unfamiliar target and a source structure that might make it intelligible. In learning, it is between past experience and future generalization. In each case, cognition acts by transforming representations so that the gap becomes smaller.

\subsection{Connection with Existing Frameworks}

Multiple existing theoretical frameworks already capture important aspects of these cognitive priors, but they were largely developed independently, each emphasizing a different facet of cognition. Free-energy minimization formalizes prediction and belief updating \citep{friston2010freeenergy, raoballard1999}; minimum description length (MDL) and algorithmic information theory (AIT) formalize compression and abstraction \citep{rissanen1978, grunwald2007, solomonoff1964, kolmogorov1965}; structure-mapping theory (SMT) formalizes analogy \citep{gentner1983structure, falkenhainer1989}; and reinforcement learning (RL) formalizes action selection under goals and rewards \citep{suttonbarto2018}. Each provides an effective account of a particular cognitive operation under a particular cognitive prior. Yet cognition does not appear to operate as a set of disconnected modules: abstraction, analogy, prediction, planning, and concept formation interact continuously in the construction and revision of representations. The purpose of the schema in Section~\ref{sec:framework} is therefore to interpret these frameworks as special cases of a single cognitive dynamics, each corresponding to a particular choice of representation $R$, transformation $T$, target $G$, discrepancy measure $\mathcal{D}$, and regularizer $\Omega$ (the last encoding a cognitive prior). Studying them this way also clarifies what is missing from existing formalisms if the goal is open-ended intelligence.

Although each of these objectives can be written abstractly as regularized discrepancy minimization, the content of the discrepancy and the regularizer differs substantially from one framework to the next, and that difference is where the substance lies. The discrepancy $\mathcal{D}$ may be a compression residual, a prediction error, a relational mismatch, a control error, a negative return, or inconsistency in explanation. The regularizer $\Omega$ encodes a different cognitive bias in each case, for example simplicity in MDL and AIT, prior plausibility in Bayesian inference, parsimonious causes in free-energy minimization, relational analogy in structure-mapping, and goal-directed control in reinforcement learning. These priors are not arbitrary mathematical constructions: each is what makes its corresponding cognitive operation computationally tractable. We therefore read the frameworks as slices of one cognitive process, and viewing them this way makes visible what is left out: the operation that changes the representational frame itself. In the paragraphs below, we identify for each framework the corresponding cognitive operation $T$ and the prior $\Omega$, as summarized in Table~\ref{tab:framework-cognitive-priors}.

\paragraph{Compression and abstraction (MDL, AIT)}
MDL seeks a model minimizing $L(M) + L(O\mid M)$, where $L(M)$ is the cost of describing the model itself, and the residual $L(O\mid M)$ is the cost of describing the observation---what the model $M$ leaves unexplained. The cognitive operation is \emph{abstraction}, which replaces surface-level observations with a shorter, invariant structure that can be used to construct them. The cognitive prior is \emph{simplicity}, $\Omega = L(M)$, the model's own description code length \citep{rissanen1978,grunwald2007}. AIT captures the same pairing, using program length as measure of the description length. Writing $K(\cdot)$ for Kolmogorov complexity---the length of the shortest program that outputs a given string---and $h$ for a hypothesis, The preferred hypothesis is $h^\ast \in \arg\min_h[\,K(h)+K(O\mid h)\,]$, where $K(h)$ the program length of the hypothesis and $K(O \mid h)$ the residual information in the data once $h$ is supplied. The operation is again abstraction, now in a strong form as shortest-program discovery under the prior of simplicity $\Omega = K(h)$ \citep{solomonoff1964,kolmogorov1965,livitanyi2008}. Because $K$ is incomputable, this is an idealization rather than an algorithm. Its computable, resource-bounded form is exactly the $L^{B}_{\Lang}$ of Section~\ref{sec:vocab-gap}. The slice of cognition these frameworks capture is the compressive nature of concept formation: concepts formed by compressing given observations. But both fix the description basis, and neither explains how a bounded system invents the new primitives that enlarge that basis and make future descriptions shorter. This missing step is precisely the vocabulary gap.

\paragraph{Analogy (SMT)}
Structure-mapping theory models analogy as the alignment of relational structure between a source $S$ and a target $U$: a mapping $m$ places elements of $S$ in correspondence with elements of $U$ so that the \emph{relations} among them are preserved, not surface-level features. Its construction follows the \emph{systematicity} principle: among candidate mappings, those that align deep, mutually interconnected systems of relations and higher-order relations, are preferred over those resting on shared attributes \citep{gentner1983structure, gentnermarkman1997}. The structure-mapping engine realizes this by searching for maximal structurally consistent correspondences under constraints of one-to-one correspondence and parallel connectivity, scored by systematicity \citep{falkenhainer1989}. The structure mapping $m$ can be conceptually characterized by the objective
\[
m^\ast \in \arg\min_m\big[\,\mathrm{Mismatch}(m(S),U) + \alpha\,\mathrm{Cost}(m) - \beta\,\mathrm{Sys}(m)\,\big].
\]

\noindent in which $\mathrm{Cost}(m)$ is a parsimony penalty on the size of the mapping. Under the schema of Section~\ref{sec:framework}, the discrepancy $\mathcal{D}$ is the residual relational mismatch (function $\mathrm{Mismatch} + \alpha\mathrm{Cost}$ above), the cognitive operation $T$ is \emph{analogy}, and the cognitive prior $\Omega$ is systematicity: a bias toward relational coherence over surface likeness $\Omega \propto -\mathrm{Sys}(m)$.

\paragraph{Inference and belief update (free energy, Bayes)}
The free-energy principle is one of the closest existing frameworks to the cognitive discrepancy schema~\citep{friston2010freeenergy}, because it explicitly treats perception, action, and learning as the reduction of mismatch between a generative model and the world. Let $q(z)$ be an approximate posterior over latent states $z$, and let $p_R(O,z)$ be the generative model encoded by the current representational state $R$. Variational free energy can be written as

\[
\begin{aligned}
\mathcal{F}(q,R) &= \mathbb{E}_{q(z)}\big[\log q(z) - \log p_R(O,z)\big] \\
&=  \mathbb{E}_{q(z)}\big[-\log p_R(O\mid z)\big] + D_{\mathrm{KL}}\!\big(q(z)\,\|\,p_R(z)\big)
\end{aligned}
\]

\noindent The first term measures the prediction error and how poorly the model explains the observations. The second term is a complexity measure which discourages posterior beliefs $q(z)$ from moving from the prior $p_R(z)$. In the schema of Section~\ref{sec:framework}, the cognitive operators are \emph{perceptual inference} (updating $q$ to better explain observations) and \emph{active inference} (acting to change future observations $O$), and the cognitive prior is parsimony and coherence in explanation, which updates beliefs that stay coherent with prior expectation \citep{raoballard1999, friston2010freeenergy, clark2013}.

Bayesian inference is closely related but narrower. It can be written as minimizing negative log posterior probability:

\[
h^\ast \in \arg\min_h\left[-\log p(O\mid h)-\log p(h)\right],
\]

\noindent where the first term measures prediction error and the second term reflects the prior plausibility of the hypothesis (or internal state representation). Cognitively, the operation is belief revision over a hypothesis space, guided by a prior preference coherent with background knowledge \citep{tenenbaum2011}. Bayesian inference therefore captures an important form of rational belief update, but like free-energy minimization, it usually assumes that the hypothesis space is already specified. In open-ended intelligence the system must revise the hypothesis language itself.

\paragraph{Curiosity and compression progress}

Compression-progress curiosity is best understood as a trajectory-level extension of the compression frameworks above. Let $h_{\le t}=(o_1,\dots,o_t)$ denote an agent's experience up to time $t$, let $\theta_t$ parameterize its current predictive model or compressor, and let $L(h_{\le t};\theta_t)$ be the description length of the observed history under that model. This is the same quantity that MDL seeks to minimize: a better model is one that compresses the observation more effectively\citep{rissanen1978}. Updating $\theta_{t-1}$ to $\theta_t$ so that the same history becomes more compressible is an abstraction step guided by a simplicity prior. closely related to the compression view of MDL and AIT~\citep{solomonoff1964,schmidhuber2010}.

What curiosity adds is that it does not reward compression itself, but \emph{improvement in compression}. This makes curiosity different from pure MDL. MDL asks how to compress the data already given, but curiosity asks which data should be sought because that is expected to produce future compression progress. The cognitive operation $T$ therefore includes two layers: abstraction for experience compression, and exploration for learning experience maximization. 

The corresponding cognitive prior is not simplicity alone, but \emph{learnable progress}. The agent should be motivated to explore regions where its current representation is insufficient for explanation but improvable. Curiosity is therefore the framework that comes closest to valuing representational change. Its objective is defined not on discrepancy at a single state, but on the rate at which discrepancy is reduced along a trajectory.

However, curiosity still does not close the verifier gap in its strongest form. Compression progress is measured against the current experience stream, model class, and notion of what counts as compressible. It can reward a primitive for making accessible future experience more predictable, but it cannot judge whether a primitive will matter for tasks or standards of success that are not yet expressible in the present representation. That forward-looking value is precisely what the verifier gap needs, and what a fixed compression objective cannot supply.

\begin{table}[t]
\centering
\footnotesize
\renewcommand{\arraystretch}{1.35}
\newcolumntype{Y}{>{\raggedright\arraybackslash}X}
\begin{tabularx}{\textwidth}{@{}>{\raggedright\arraybackslash}p{1.9cm} Y Y >{\raggedright\arraybackslash}p{2.2cm} Y@{}}
\toprule
\textbf{Framework} & \textbf{Discrepancy $\mathcal{D}$} & \textbf{Regularizer $\Omega$} & \textbf{Operation $T$} & \textbf{Cognitive prior} \\
\midrule
MDL \citep{rissanen1978} &
Data uncompressed by model, $L(O\mid M)$ &
Model cost $L(M)$ &
Abstraction, model selection & Simplicity \\

AIT \citep{solomonoff1964} &
Residual after hypothesis, $K(O\mid h)$ &
Program length $K(h)$ &
Abstraction, shortest-program search &
Simplicity \\

SMT \citep{gentner1983structure} &
Relational mismatch &
Systematicity $-\beta\,\mathrm{Sys}(m)$ &
Analogy &
Systematicity \\

Free energy\citep{friston2010freeenergy} &
Expected surprise &
Complexity $D_{\mathrm{KL}}(q(z)\,\|\,p(z))$ &
Active inference &
Coherence, parsimony \\

Bayesian inference \citep{tenenbaum2011} &
Prediction error &
Prior cost $-\log p(h)$ &
Belief revision &
Coherence \\


Curiosity \citep{schmidhuber2010} &
Uncompressed experience &
Compression &
Exploration, abstraction &
Exploration and learning \\

\bottomrule
\end{tabularx}
\caption{
Established frameworks read as restricted forms of cognitive discrepancy reduction, each fixing a discrepancy $\mathcal{D}$ and committing to a cognitive operation $T$ guided by a prior $\Omega$. Each captures one slice of cognition.
}
\label{tab:framework-cognitive-priors}
\end{table}
 
\subsection{What the Comparison Establishes}

Each framework above can be interpreted as a specific instantiation of the discrepancy-reduction schema: a particular cognitive operation, guided and regulated by a corresponding cognitive prior that reduces a particular kind of discrepancy. Simplicity incentivizes abstraction, systematicity incentivizes analogy, curiosity incentivizes exploration, and so on. Each theory models one slice of cognition rather than cognition as a whole, but interpreting them under the same discrepancy-reduction schema helps consolidate these slices as facets of a common process rather than as disconnected modules. Seen together, though, they share the omission this paper is concerned with: every operation they formalize searches \emph{within} a fixed representation $R$, and none changes the representation frame $R$ itself:
\[
R_{t+1} = T_t(R_t), \qquad T_t \in \big\{\,\text{association, abstraction, analogy, composition,}
\]
\vspace{-1.6em}
\[
\qquad\qquad\qquad\qquad \text{concept invention, evaluator revision}\,\big\}
\]

Open-ended intelligence is not a faculty apart from perception, inference, analogy, and reasoning. It is a continuation of these activities at the point where intra-space transformations are no longer sufficient. At present no framework above formalizes the inventive, self-evolving cognition that generates new primitives and revises its own criteria of evaluation. That missing piece is what the two gaps name. The vocabulary gap is whether a system can generate such frame-changing transformations at all, and the verifier gap is whether it can evaluate them when the prior $\Omega$, the discrepancy $\mathcal{D}$, and the target $G$ are themselves changing.

Addressing these problems is not just about how to make open-ended intelligence possible, but ultimately a question of how to build future AI systems and make them more capable. If current systems are trained primarily through next-token prediction, then they are implicitly optimizing narrow slices of the discrepancy-reduction schema. A broader path may require training objectives, data formats, memory structures, and evaluation designs that more directly cultivate cognitive operations, which we discuss in Section~\ref{sec:improvement}.

\section{Levels of Innovation Autonomy}
\label{sec:levels}

Recent progress in AI for scientific and mathematical discovery has been rapid. A growing family of systems, under the category of ``AI co-scientist'', now generate research papers, propose new theorems, discover new algorithms, and report novel findings autonomously \citep{lu2024aiscientist, gottweis2025coscientist, romera2024funsearch, novikov2025alphaevolve}. However, what is largely missing is a clear account of the nature of this novelty: how hard the underlying creative step was, and how far can it go beyond searching within a fixed representation frame. A system that finds an efficient sorting algorithm and a system that introduces a new concept around which a field gets reorganized are both labeled as ``discovery'', yet they are not the same kind of act. To see where current systems actually stand, how innovative they are, and how far they remain from true open-endedness \citep{pmlr-v235-hughes24a}, it helps to define a ladder of innovation autonomy and to place systems on it with explicit criteria.

The ladder is organized in three dimensions: 1) Search pattern: is the system searching within a fixed representational space, or can it modify the space being searched?  2) Vocabulary autonomy: can it create and reuse new conceptual primitives? 3) Verifier autonomy: does it own the verifier and evolve it as needed, or is the success criteria supplied and fixed from outside? Accordingly, we define four levels of innovation autonomy, summarized in Table~\ref{tab:ladder-autonomy}.

\paragraph{L0: Direct intra-space search}
The system searches within a fixed representational space $R$. The task, vocabulary, tools, and success criterion are supplied and fixed. The system generates a plan, solution, explanation, program etc., but it does not autonomously run an improvement loop. Standard LLM assistants, retrieval-augmented systems (RAG)~\citep{lewis2021retrievalaugmentedgenerationknowledgeintensivenlp}, chain-of-thought prompting, and tool-use agents belong here when they provide responses to tasks without verifier feedback.

\paragraph{L1: Intra-space search loop with fixed verifier}
L1 systems still search within a fixed representation, but they are equipped with an autonomous improvement loop. They generate candidates, evaluate them with a fixed verifier, make improvements based on feedback, and repeat. FunSearch~\citep{romera2024funsearch}, AlphaEvolve-style~\citep{novikov2025alphaevolve} systems belong here. They can produce novel discoveries due to powerful self-improvement loop, but the representation space, language, objective, and verifier remain fixed from outside. L1 therefore improves search autonomy, not vocabulary or verifier autonomy.

\paragraph{L2: Scaffolded representation-space augmentation}
The system also has an improvement loop but searches over a family of an augmented representational spaces, abstractions, hypotheses, research plans, experiments, or libraries. DreamCoder~\citep{ellis2021dreamcoder} and Stitch-style~\citep{bowers2023stitch} systems fit here because they learn reusable library abstractions that change the local language of future program search. AI-for-science system such as The AI Scientist and AI co-scientist-style architectures, as well as autonomous lab agents like~\cite{biomni2026science,Panapitiya2026AutoLabs} also fit here as they search across research ideas, experimental designs, code implementations, and paper drafts. In both cases, the system explores more than one fixed candidate space.

However, L2 remains scaffolded. The task domain and family, tool stack, literature corpus, simulator, and evaluation criterion is supplied externally. The system may learn local abstractions or propose novel hypotheses, but it does not fully decide which problem should exist or how new primitives should be valued beyond the given frame. L2 therefore partially addresses the vocabulary gap but does not narrow the verifier gap.

\paragraph{L3: Open-ended innovation autonomy}
At L3, the system is capable of changing not only candidates within a frame, but the representational and evaluative frame itself. It can detect persistent cognitive discrepancies, decide which gaps are worth pursuing, invent new conceptual primitives, stabilize them in memory, reuse them across contexts, and revise its evaluative criteria with the expansion of the representation space. This is a generative frame change, it requires closing both the vocabulary and verifier gap. No current general AI system clearly reaches this level.

\paragraph{Where current systems stand}

\begin{figure}[h]
    \centering
    \includegraphics[width=0.65\linewidth]{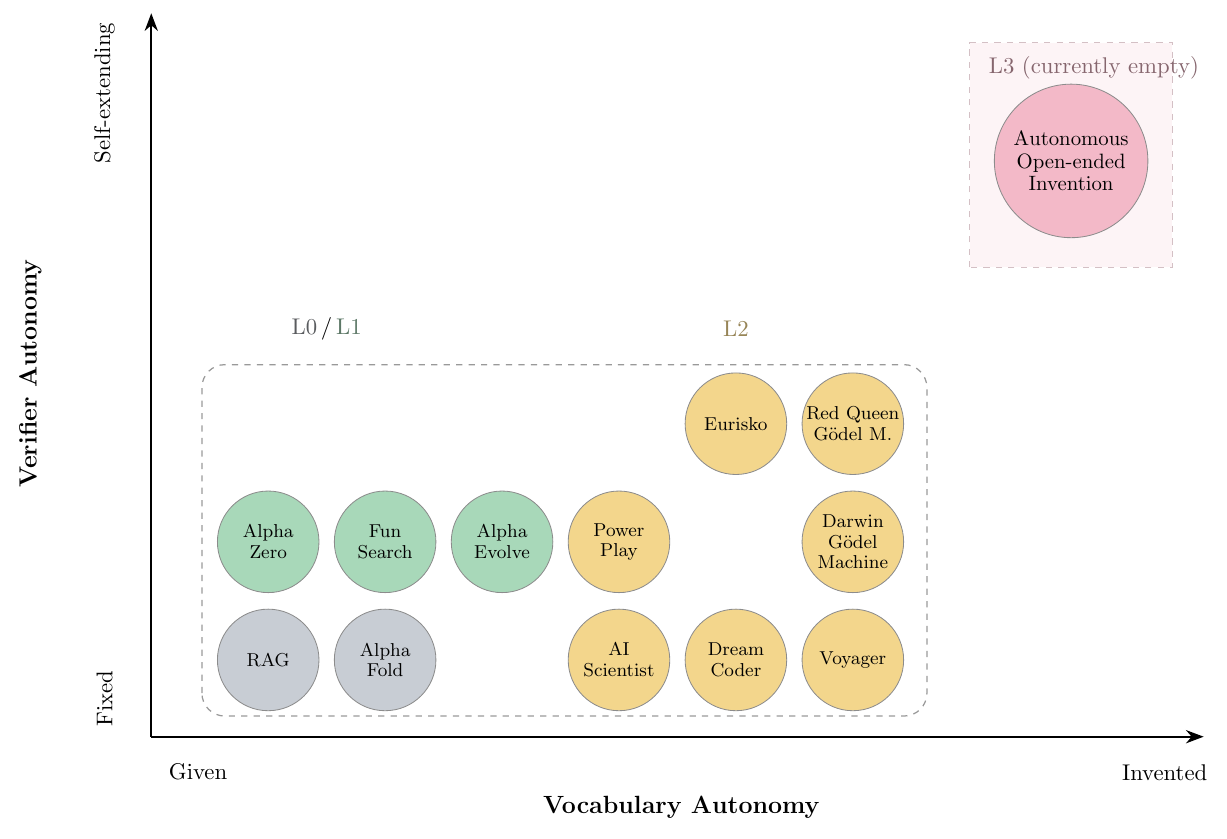}
    \caption{{\textbf{Axes of open-ended intelligence.}}
    Systems can be positioned by Axis X (vocabulary autonomy, from given to invented) and Axis Y (verifier autonomy, from fixed to self-extending). L0 and L1 remain near the bottom-left corner, differing mainly in search autonomy, L2 moves rightward by introducing limited autonomy over primitives, L3 occupies the upper-right region, corresponding to truly open-ended intelligence where both vocabularies and evaluators become self-extending. The systems considered here include RAG~\citep{lewis2021retrievalaugmentedgenerationknowledgeintensivenlp}, AlphaFold~\citep{jumper2021alphafold}, AlphaZero~\citep{silver2018alphazero}, AlphaEvolve~\citep{novikov2025alphaevolve}, Voyager~\citep{wang2023voyager}, DreamCoder~\citep{ellis2021dreamcoder}, PowerPlay~\citep{schmidhuber2013powerplay}, Eurisko~\citep{lenat1983eurisko}, Darwin G\"odel machine~\citep{zhang2026darwingodelmachineopenended}, Red Queen Darwin G\"odel machine~\citep{iacob2026redqueengodelmachine}. Current AI systems largely cluster along the bottom edge: while rapid progress has been made in effective solution search, the spaces of primitives and evaluators are mostly fixed. The L3 region is still empty.}
    \label{fig:autonomy-position}
\end{figure}

Based on the autonomy ladder defined above, Figure~\ref{fig:autonomy-position} provides a conceptual map of representative existing systems, positioning them along two axes of vocabulary autonomy and evaluator autonomy. This ladder clarifies why current AI-for-discovery systems are impressive but still distant from open-ended innovation autonomy. The headline results new algorithms---competition-level theorems, new scientific discoveries---come overwhelmingly from lower rungs (L0, L1) of the ladder, and concentrate in the domains where primitives for search are fixed, and verification is cheap and automatable. The novelty they produce is bounded by the human-supplied framework that structures the problem space and verification. The top rung is empty.
 
What current systems do as strong search and compositional discovery, is undeniably real and useful. But characterized by the dimensions of the ladder, almost all of it is intra-space search under human-framed automation. Very little of it touches the operations that change the frame. What separates these systems from open-endedness is therefore not model size, data quantity or inference computation scaling, context length, or agentic scaffolding. Those proven techniques advance a system along the search and scope dimensions, but does not necessarily narrow the distance from open-endedness. To address these outstanding issues to achieving stronger intelligence, we will need to rethink about the way AI systems are built, because there is a qualitative difference between building closed- and open-ended systems. We discuss in the next section.

\begin{table}[t]
\centering
\small
\begin{tabular}{p{0.07\linewidth}p{0.22\linewidth}p{0.22\linewidth}p{0.24\linewidth}p{0.18\linewidth}}
\toprule
Level & Vocabulary autonomy & Verifier autonomy & Search pattern & Representative systems \\
\midrule

L0 &
Fixed representation space and vocabulary &
No autonomous verifier loop. Evaluation is external or human-directed &
Direct intra-space search, exploration inside a fixed representation &
Standard LLM assistants, RAG, chain-of-thought prompting, basic tool-use agents \\
\hline
L1 &
Fixed representation and vocabulary, proposals improve but primitives do not change &
Fixed external verifier providing feedback during iteration &
Iterative intra-space search, generate $\rightarrow$ evaluate $\rightarrow$ select $\rightarrow$ improve candidates within a fixed space &
FunSearch, AlphaEvolve-style systems, coding agents with unit tests, theorem-proving loops with checkers \\
\hline
L2 &
Partial vocabulary autonomy. System can learn or extract local invariants as abstractions, or search in extended but scaffolded research space &
Verifier remains externally scaffolded by benchmarks, task distributions, simulations, or human objectives &
Scaffolded representation-space search: search over local spaces or frame variants under augmented but fixed frame &
DreamCoder, Stitch-style library learning, AI co-scientist systems, autonomous lab agents \\
\hline
L3 &
High vocabulary autonomy. System can invent, stabilize, store, and reuse new primitives across domains &
High verifier autonomy. System can construct and revise standards for new primitives valuation, and create new experimental setups &
Generative frame change: autonomous gap identification, representation expansion, persistent reuse, and evaluator revision &
No clear current AI system example \\

\bottomrule
\end{tabular}
\caption{A ladder of innovation autonomy.}
\label{tab:ladder-autonomy}
\end{table}

\section{What Would Have to Change}
\label{sec:improvement}

If the vocabulary and verifier gaps are central, then the path to open-ended AI cannot be reduced to scaling models, extending context windows, adding tools, or increasing inference-time search. These techniques improve search within an existing frame, but they do not address how a system should recognize that the frame itself is inadequate. The deeper challenge is to build systems that can generate, stabilize, test, revise, and reuse new primitives under uncertainty about their future value, and in the hardest cases, construct the very criteria by which that value can be judged. Several directions follow.

\paragraph{Objectives that reward useful representation change}
\label{sec:objectives}

Next-token prediction (NTP) can be interpreted as a specific instance of the discrepancy-reduction schema in Section~\ref{sec:framework}: the model is trained to reduce the discrepancy between a context representation and the next-token distribution, typically measured by cross-entropy. In this case $R$ is the model's representation of the token context, $G$ is the observed next token as a target distribution over the fixed vocabulary, $\mathcal{D}$ is the cross-entropy loss, and the transformation $T$ is identity $T=I$ since there is no representation frame change. This objective has been remarkably effective, but it rewards abstraction only indirectly. A representation is favored only insofar as it improves prediction over the training distribution. The model is not explicitly trained to notice that its current vocabulary is inadequate, introduce a new abstraction, and preserve that abstraction for future reuse.

The objective should therefore be expanded from answer prediction to representation revision. A training episode should not only ask which output is correct, but also whether the current representation leaves a recurring discrepancy unresolved. A system should receive credit for triggering an operation $T$ that produces a more useful representation $R' = T(R)$ for solving a problem, including abstraction, compression, relational mapping, composition etc.

For example, the current state $R$ may contain a collection of observations that appear heterogeneous under the existing vocabulary. The target $G$ is not a single next token, but a representation that captures their shared structure. The discrepancy $\mathcal{D}$ then measures how poorly the current representation explains, compresses, or transfers across the observations. A useful transformation is one that reduces this discrepancy by introducing an abstraction that can be reused in later cases.

Under the general schema of Eq.\ref{eq:discrepancy}, richer objectives can therefore be constructed by varying the representation $R$, the transformation $T$, and the discrepancy measure $\mathcal{D}$. The goal is not to replace prediction, but to augment it with explicit reward for useful representational change. This would make concept formation and abstraction direct targets of learning rather than accidental byproducts of NTP.

\paragraph{Data that elucidates invention, not only its outcome}
\label{sec:data}

Enrichment of objectives requires corresponding change in data. Almost all existing training data exhibits concept \emph{use} rather than concept \emph{invention}. Mathematics already uses the established concept of ``eigenvalue'', physics already uses the concept ``entropy''. A distribution that only displays established frames is a weak teacher of frame change, because the crucial act of novel primitive introduction is usually absent.

Data for open-ended intelligence should instead expose trajectories of representational revision. A useful trajectory would show an initial representation $R_t$, a set of observations or tasks that produce a persistent discrepancy under that representation, a transformation $T_t$ that introduces a candidate primitive, and a revised representation $R_{t+1}$ in which the discrepancy is reduced. The training target is not merely the final answer, but the operations that made it possible to represent the answer.

This suggests a different data format from the ordinary NTP over finished text. Training examples should introduce explicit structures before and after concept invention: what system could express before the primitive, what discrepancy remained unresolved, what primitive was introduced, and how later tasks were solved once the new primitive became available. Such data would not solve open-ended intelligence by itself, but it would better equip models to recognize and perform frame-changing actions, rather than merely absorbing the surface forms of already stabilized knowledge. For example, a mathematical trajectory could show several problems that are hard to interpret under the
original concept set, followed by the introduction of a definition that turns them into instances of reusable patterns. A scientific trajectory could show observations that resist explanation until a new causal relation is introduced. A programming trajectory might show repeated code patterns being compressed into a reusable library, after which later programs become shorter and easier to synthesize. In each case, the important feature is not simply that a new primitive appears, but that it is later reused to reduce description length and search cost.

Under the discrepancy-reduction schema, this means that data should instantiate diverse choices of $R$, $T$, and $\mathcal{D}$, including representations that fail in different ways, transformations augment those representations, and discrepancy measures that reveal why the revision matters. The goal is to train models not only to use established concepts, but to recognize when a new primitive is needed to make future prediction feasible or search more efficient.

\paragraph{Persistent primitive stores and consolidation}
\label{sec:stores}

A newly proposed primitive is useful only if it persists long enough to be reused. This is why in-context learning is not enough. A model may temporarily invent a helpful abstraction inside a single context window, but once the context disappears, the abstraction does not become part of the system's cumulative representational resources. Open-ended intelligent systems therefore need selective stores to keep candidate primitives. The form of the primitive will depend on the domain. In program synthesis, it may be a reusable function or library abstraction. In science, it may be a new variable or causal relation. In mathematics, it may be a definition, or proof pattern. What matters is not the specific storage format, but whether the primitive can be retrieved, tested, and reused in future tasks.

Persistent memory, however, must remain revisable. Without periodic pruning, a primitive store
can degenerate into a collection of redundant or noisy abstractions. A primitive should therefore retain its place in durable memory only if it continues to demonstrate value across future cases. The criteria of amortized compression and feasibility extension introduced in Section~\ref{sec:vocab-gap} provide exactly this test: a primitive is worth preserving only if its presence reduces solution cost across a task family, if removing it harms performance, or if it makes tasks feasible that were previously unreachable under bounded search budgets.

A primitive that proves repeatedly useful should then be internalized into a model. It is no longer retrieved as an external tool or memory entry, but injected into the system's intrinsic knowledge. This is analogous to the wake-sleep consolidation in DreamCoder\citep{ellis2021dreamcoder}, where abstractions discovered from solved programs are used to update the models so that their future problem solving ability can be improved. Technically, consolidation could be realized through fine-tuning or continual learning, but conceptually, the
important requirement is that open-ended learning must contain a loop for converting
successful inventions into stable future vocabulary and permanent representations once their usefulness has been demonstrated.

\paragraph{Surrogate verifiers}
\label{sec:surrogate}

A big challenge in open-ended intelligence is narrowing the verifier gap. One form of this gap, discussed in Section~\ref{sec:verifier-gap}, arise from delayed and expensive verification. Open-ended discovery is long-horizon, and its outcome verifier is frequently unavailable at the moment a candidate must be judged. The effectiveness of a newly designed drug, the correctness of a new physical theory, or the true cause of a manufacturing anomaly cannot be scored immediately as a definitive test may take months or even years. A system solving these problems cannot let outcome verification guide search. It must use a surrogate verifier, a cheap and computable estimate of a candidate's value, to identify promising search directions.

How a surrogate can be built depends on what evidence is available at hand. If historical data of prior trials, prior proofs, prior anomaly resolutions is available, then surrogate could possibly be learned as a value model that predicts eventual outcome from early or partial evidence. However in open-ended discovery cases, such data may not be easily available. In such cases, the surrogate needs to fall back on the cognitive priors described in Section~\ref{sec:framework}. Each prior $\Omega$ that makes a cognitive operation tractable also serves as a value signal in the absence of outcome feedback. For example, description length reduction of the data a hypothesis can explains (compression), the rate of description length reduction (compression progress~\citep{schmidhuber2010}), alignment of deep relational structure over surface features (systematicity), or predictive gain on held-out observations. In general, the regularizers of Table~\ref{tab:framework-cognitive-priors} can be repurposed for provisional value estimation.

It should be noted that in both cases the surrogate is a heuristic, not the objective. Optimizing and overfitting on the surrogate directly could lead to degenerate solutions. A system rewarded for satisfying its surrogate will find ways to satisfy it without maximizing the underlying value, as Eurisko's self-crediting heuristics did~\citep{lenat1983eurisko}. A usable surrogate must therefore be recalibrated against real outcomes whenever they arrive, and overridden when they disagree.

\paragraph{Self-extending evaluation}

Another form of the verifier gap, introduced in Section~\ref{sec:verifier-gap}, arises from the fluid nature of verification in open-ended intelligence. A system that expands its own
representation space and vocabulary must also evolve the criteria by which it judges its generation, because a fixed evaluator cannot score objects it was never designed to represent. This is the crucial part, on the evaluation side, of the persistent primitive store proposed above. For a primitive store to retain genuinely novel primitives, the system cannot rely on a static verifier, it must innovate evaluation procedures as the search progresses. For example, a newly abstracted function module may require unit tests be synthesized on the fly. A newly proposed scientific hypothesis may require intermediate experiments designed to test it. An invented concept requires subsequent usage cases to reveal whether it captures a genuine invariant across broad contexts rather than a local abstraction, and correspondingly, whether it reduces description length across wider range of tasks rather than only for the case at hand.

Vocabulary invention and evaluator extension are therefore two aspects of the same operation. A system cannot meaningfully expand its representational space unless it also expands the methods by which newly introduced objects can be tested and compared. Representational expansion becomes feasible only when what was previously unexpressible or untestable is made verifiable. This is the capability most current systems lack. They can usually optimize effectively within a fixed evaluative frame, and some can introduce local abstractions under a fixed task distribution, but they do not yet possess a principled mechanism for deciding when the evaluation frame itself should change.

\section{Related Work}

This paper connects four lines of work that are often discussed separately: open-endedness and its crucial role in intelligence, cognitive foundations of innovation, self-modifying discovery systems, recursive self-improvement and automated research agents. The common thread is the question of whether an intelligent system can move beyond solving tasks inside a fixed representation frame and instead revise the primitives, search space, and evaluative criteria that make meaningful discovery possible.

\subsection{Open-Endedness: Necessity and Limits}
Open-ended innovation is what allowed humans to create knowledge, make discoveries, and build civilization. Modeling this capacity is substantially harder than modeling closed-ended intelligence, where the task, vocabulary, objective, and verifier are fixed in advance.
\citep{BODEN1998347} argues that creativity is a core feature of intelligence and an
unavoidable challenge for AI, grounded in cognitive capacities such as association, analogical
thinking, perception, search, and reflective self-criticism---the operations that we formalize as
discrepancy-reducing transformations in Section~\ref{sec:framework}.

Recent position work sharpens this necessity claim. Hughes~\citep{pmlr-v235-hughes24a} argue that open-endedness is essential for artificial superintelligence, defining it behaviorally through the novelty and learnability of a system's output stream relative to an observer.
Genewein et al.~\citep{genewein2026agiasi} argues that if intelligence is viewed as search, then a suitable open-ended search process yields progressively stronger general performance, identifying superintelligence with super-creativity. These framings are complementary to ours, differing in level of analysis: behavioral definitions ask whether a system produces a continuing stream of novel, learnable artifacts, whereas this paper asks what internal operation produces such behaviors---representation-space expansion, constrained by the verifier gap.
Avestimehr et al.~\citep{avestimehr2026novafundamentallimitsknowledge} provide a complementary limitation result: if retraining is support-preserving, an autonomous system cannot discover valid artifacts outside its initial generative support, so breaking exploration barriers, valid verification, and human amplification through guidance are each necessary for strong knowledge discovery. This is a formal complement to our informal diagnosis---their support barrier is closely related to the vocabulary gap, and their verification condition to the verifier gap.

\subsection{Cognitive Foundations of Innovation}
Three operations underlie the production of novelty: abstraction, analogy, and composition. In our framework, these operations are treated as discrepancy-reducing transformations over representational space (Section~\ref{sec:framework}). We situate the relevant prior work below.

\paragraph{Abstraction} Abstraction is perhaps the most basic cognitive operation behind open-ended and general intelligence, because it converts recurring structure into reusable primitives. Classical accounts treat abstraction not merely as the removal of details, but as the construction of new mental structures. Immanuel Kant~\citep{kant1781}, for example, already treats concepts not as summaries of experience but as rules of synthesis that make structured experience possible in the first place, that a new primitive does not merely compress known patterns but makes previously unavailable ones thinkable. More recently, Lawrence Barsalou\citep{barsalou2005situating} argues that abstract concepts are essential for sophisticated situated action, as they allow humans to integrate complex situational elements to comprehend events and predict outcomes.

Formal accounts give this intuition an operational interpretation. Minimum description length and algorithmic information theory frame a good abstraction as one that shortens the joint description of a family of observations, models, or solutions~\citep{solomonoff1964,kolmogorov1965}. Legg and Hutter~\citep{legg2007universal} formalize intelligence as expected reward across environments, where compression of environmental structure is central to generalization. Chollet's ARC~\citep{chollet2019measure} brings this concern into AI evaluation by emphasizing skill-acquisition under novelty, where abstraction and recombination are central. The abstraction operation this paper emphasizes is the one that changes basis and invents
primitives in which future observations become compressible at all.

\paragraph{Analogy} The classic work~\citep{GICK19831} shows that people transfer solutions more effectively when they induce an abstract schema from multiple concrete examples, providing evidence that transfer is mediated by abstraction rather than surface similarity. Mitchell~\citep{Mitchell_2021} frames conceptual abstraction and analogy as central to human learning and robust adaptation, while arguing that contemporary AI still falls short of human-like abstraction. The broader tradition treats analogy as a pillar of cognition \citep{gentner2001analogical,hofstadter2013surfaces,lakoff1980metaphors}. We adopt this view and in Section~\ref{sec:framework}, read Gentner's structure-mapping as one instantiation of discrepancy reduction under a systematicity prior.

\paragraph{Composition} Much innovation proceeds by recombination. Arthur~\citep{arthur2009nature} argues that technological innovation is largely the recombination of existing components, principles, and practices, guided by human purpose and enabled by the harnessing of natural phenomena. \citep{momennejad2026compositionalframeworkopenendedintelligence} give a compositional definition of open-endedness as infinite sequences generated by combining representational and algorithmic primitives, and propose ``next-primitive prediction'' as a stronger inference paradigm. \citep{schapiro2026combinatorialcreativitynewfrontier} argue that combinatorial creativity is a distinct form of generalization that fixed-target benchmarks miss, and should be scored by degrees of novelty and utility rather than against a fixed answer---closely aligned with our argument. \citep{artiles2026alienspacesciencesampling} operationalize this for scientific ideation by sampling atomic ideas and composing them into proposals scored for novelty and plausibility by trained judge models. These results matter, but composition operates within a fixed conceptual space as it presupposes that the units to be combined already exist. True open-ended intelligence requires more than composition, but also inventing primitives in the first place, to narrow the vocabulary gap.

\paragraph{Limits of current systems} Several recent critiques sharpen the difficulty of this problem. For example, Lerchner~\citep{lerchner2026abstraction} argues that computational systems often presuppose the abstractions, symbolizations, and semantic grounding that they are claimed to generate. Zahavy~\citep{zahavy2026jump} argue that LLMs struggle with abductive ``jumps'' of the kind required for major scientific reframing, such as the conceptual leap from Newtonian mechanics to general relativity. Whether or not one accepts these stronger critiques, they point to the same underlying issue: open-ended intelligence requires more than interpolation inside an inherited conceptual vocabulary.

\subsection{Self-Modifying Discovery Systems}
Early AI discovery systems already confronted the problem of how an automated system could generate concepts rather than merely apply them. Lenat's Automated Mathematician (AM) \citep{lenat1976}, searched for interesting mathematical concepts by mutating small Lisp programs under a body of heuristics. Its successor Eurisko \citep{lenat1983eurisko} went further, applying heuristics to revise its own heuristics, including those judging what was worth pursuing. Eurisko is, in our terms, an explicit early reach toward a self-extending evaluator, and its failure modes are instructive: some heuristics behaved parasitically, capturing the very criterion that scored them. \citep{lenat1984am}'s later reflection is equally important---much of AM's apparent conceptual fertility was an artifact of an encoding in which random program mutations happened to correspond to meaningful mathematical objects, so the apparent representation expansion was partly a property of the representation, not an autonomous achievement of the system. This is why addressing the verifier gap requires an operational, ablation-based criterion rather than an observer's impression that ``something new emerged.''

Schmidhuber's PowerPlay~\citep{schmidhuber2013powerplay} offers a more formal approach. PowerPlay continually searches for the simplest still-unsolved task together with a solver modification that solves it while preserving prior skills. Hence, it constructs a self-generated curriculum with an explicit criterion for novelty and validity. Clune's AI-GAs framework~\citep{clune2019aigas} broadens this vision by proposing systems that learn not only solutions but also architectures, learning algorithms, and environments. POET~\citep{wang2019poet} and Enhanced POET~\citep{wang2020enhancedpoet} similarly generate an ecosystem of paired problems and solutions, allowing new challenges and capabilities to co-evolve through stepping-stone transfer.

These systems are important precursors because they reject the fixed-task assumption. They show that progress can arise from a growing population of tasks, agents, and stepping stones rather than from optimization toward one predefined goal. However, their criteria for novelty, solvability, interestingness, or transfer are still largely specified by the designer. They expand the task distribution, but they do not fully solve the verifier gap. In the terminology of this paper, they make progress toward open-ended search, but they do not yet provide a general mechanism by which the system invents and validates the evaluative criteria under which future primitives are judged.

\subsection{Recursive Self-Improvement and Automated Research}
Recursive self-improvement converts search experience into better search machinery, through model weights or external harnesses. \citep{schmidhuber2006goedelmachinesselfreferentialuniversal}'s
G\"odel Machine (GM) is the classical statement: a self-referential solver that rewrites any part of its own code, including the proof searcher and the self-improvement mechanism itself, once it proves the rewrite beneficial under an encoded utility function. Modern successors retain the self-modifying principle but replace proof with empirical or learned improvement, including the Darwin GM \citep{zhang2026darwingodelmachineopenended}, Huxley GM
\citep{wang2025huxleygodelmachinehumanlevelcoding}, and Red Queen GM
\citep{iacob2026redqueengodelmachine}. Self-modification is central to all of them, but none explicitly targets the vocabulary gap. The Red Queen machine is a partial exception on the verifier axis, co-evolving the verifier with the solver during search.

Another line of work in automated research and algorithm discovery couples foundation models with search and external evaluation. FunSearch~\citep{romera2024funsearch} shows that LLM-generated programs, when embedded in an evolutionary loop with an automatic evaluator, can discover useful mathematical constructions. AlphaEvolve~\citep{novikov2025alphaevolve} extends this paradigm to broader algorithmic discovery and optimization. EvoX~\citep{liu2026evoxmetaevolutionautomateddiscovery} moves one level higher by evolving not only candidate programs but also aspects of the search strategy used to generate them. AI Scientist and AI co-scientist systems~\citep{lu2024aiscientist,gottweis2025coscientist} generate, critique, and refine research hypotheses under researcher-provided objectives and validation procedures. Surveys of AutoResearch AI~\citep{tie2026autoresearchaiaipoweredresearch} describe this emerging pattern as an agentic loop in which generation, evaluation, refinement, and tool use are repeatedly applied to research tasks. These systems demonstrates that when the designer supplies a reliable evaluator, a clear objective, and a structured search loop, AI systems can make nontrivial discoveries. Their success therefore supports the centrality of verification. But it also highlights the limitation emphasized in this paper. The current generation of automated research systems is most effective where the target representation, evaluation protocol, and success criterion are already available. They can optimize within a supplied frame, and in some cases improve the search machinery itself, but they do not yet autonomously decide when the representation frame or evaluation frame should change.

\subsection{Conceptual Spaces and Representation Transformation}
G\"ardenfors conceptual spaces framework~\citep{gardenfors2000conceptual} models concepts as regions in geometries defined over quality dimensions. This provides a useful language for thinking about concepts not as isolated symbols but as structured regions within a representational space. Gentner's structure-mapping theory~\citep{gentner1983structure} describes analogy as alignment between relational structures.

From these perspectives, the cognitive operations discussed in this paper can be viewed as transformations over conceptual space. Association moves attention between nearby or historically linked regions. Analogy maps relational structure from one region or domain into another. Abstraction compresses multiple cases into a reusable higher-order region or primitive. Composition combines existing primitives into new configurations. Self-extending evaluation changes the criteria by which regions and transformations are judged. The central claim is therefore not simply that intelligence searches a conceptual space, but that open-ended intelligence modifies the space in which future search occurs.

\section{Limitations and Open Problems}

The purpose of this paper is to clarify the gap between current AI systems and stronger forms of AI capable of open-ended intelligence. We have proposed a framework for describing this gap in terms of representational expansion, primitive formation, delayed verification, and self-extending evaluation. However, the framework remains primarily diagnostic and conceptual. Its formal definitions and architectural implications require further development. In particular, future work should sharpen the proposed notions of vocabulary and verifier gaps, primitive usefulness, and innovation autonomy through more precise formalization and controlled experiments.

A second limitation is that human innovation is a complex cognitive process involving many interacting mechanisms, including abstraction, analogy, composition, causal modeling, memory, exploration etc. This paper does not claim to provide a complete theory of human innovation or scientific discovery. Rather, it identifies several computationally relevant mechanisms that are underdeveloped in current AI systems. An important open problem is to determine which aspects of human innovation are computationally tractable, which can be modeled algorithmically, and which may require fundamentally different modeling assumptions. It remains possible that some dimensions of human creativity depend on faculties such as grounding, embodiment, consciousness that are not fully captured by the computational procedures considered here.

A third limitation is that self-extending evaluation can raise significant safety concerns. A system that can revise its own evaluators may also drift from its original objectives, game its surrogate signals, or create harmful primitives. Capability improvement and safety analysis therefore must proceed together, not in sequence. The same mechanism that would make an L3 system open-ended is also what would make it difficult to constrain. Any practical L3 architecture would need human oversight, interpretability, external validation, controllability, and limits on self-modification built in from the start. We therefore should regard the design of such constraints as an open problem of equal criticality to the capability development itself.

\section{Conclusion}

Human intelligence is defined not only by task performance but by innovation. Civilization is, in large part, an unbounded recursion in what can be represented, constructed, and transformed. Every scientific theory, medical treatment, transportation system, and social institution demonstrates that human intelligence does not merely solve predefined problems but creates problem spaces that did not previously exist. This capacity, to reshape environments, generate new concepts, and expand the space of what can be known and built, is largely absent from today's AI systems.

While rapid progress on intellectually demanding tasks such as coding and mathematics has been made, these are mainly advances within fixed representational frames. Higher forms of open-ended intelligence require something further, so that a system can refine and expand its representational space on its own. This paper has argued that two gaps stand in the way. Closing the vocabulary gap would let a system abstract observations into novel primitives and stabilize them for future reuse, and closing the verifier gap would let a system move beyond a fixed, externally supplied standard of success, so that whether a representational expansion is worth keeping can be judged adaptively rather than declared in advance.

To characterize these gaps more precisely, we placed innovation alongside other cognitive behaviors like perception, reasoning, exploration under a unified schema of intelligence as cognitive discrepancy reduction, where intelligent behaviors are treated as a sequence of representational transformations guided by a small set of priors favoring simplicity, regularity, and causal coherence. From this perspective, established frameworks such as compression, analogy, and prediction each formalize one slice of that process, yet none addresses the operation that changes the representational and verification frame itself. This diagnosis also points toward potential remedies, for example deriving training objectives directly from the discrepancy-reduction schema rather than from next-token prediction alone, data formatting and curricula that make representational change necessary, architectures with persistent memory for invented primitives, and verification protocols that can extend themselves.

The trajectory of AI development cannot remain confined within a fixed representational closure. The goal should be a system that continually identifies discrepancies and expands itself to resolve them, rather than one frozen at a fixed capability checkpoint and agentic harness. Open-ended intelligence, as a higher form of intelligence, subsumes the closed-ended systems we have today, and the one that will require most of the effort ahead.

\bibliographystyle{plainnat}
\bibliography{references}
\end{document}